**Title:** Unsupervised Abnormality Detection through Mixed Structure Regularization (MSR) in Deep Sparse Autoencoders.

**Authors:**

Moti Freiman* PHD, Ravindra Manjeshwar PHD, and Liran Goshen PHD

*Corresponding Author (Address for correspondence)

Moti Freiman, Philips Medical Systems Technologies Ltd.

E-mail: moti.freiman@philips.com

**Funding information:** None.

**Article type:** Original research

Accepted for publication in "Medical Physics"

**Abstract:**

**Purpose:** The purpose of this study is to introduce and evaluate the mixed structure regularization (MSR) approach for a deep sparse autoencoder aimed at unsupervised abnormality detection in medical images. Unsupervised abnormality detection based on identifying outliers using deep sparse autoencoders is a very appealing approach for computer aided detection systems as it requires only healthy data for training rather than expert annotated abnormality. However, regularization is required to avoid over-fitting of the network to the training data.

**Methods:** We used Coronary Computed tomography Angiography (CCTA) datasets of 90 subjects with expert annotated centerlines.

We segmented coronary lumen and wall using an automatic algorithm with manual corrections where required. We defined normal coronary cross-section as cross-sections with a ratio between lumen and wall areas larger than 0.8. We divided the datasets into training, validation, and testing groups in a 10-fold cross validation scheme. We trained a deep sparse overcomplete autoencoder model for normality modeling with random structure and noise augmentation. We assessed the performance of our deep sparse autoencoder with mixed structure regularization without denoising (SAE-MSR) and with denoising (SDAE-MSR) in comparison to deep sparse autoencoder (SAE), and deep sparse denoising autoencoder (SDAE) models in the task of detecting coronary artery disease from CCTA data on the test group.

**Results:**

The SDAE-MSR achieved the best aggregated Area Under the Curve (AUC) with a 20% improvement and the best aggregated Average Precision (AP) with a 30% improvement upon the SAE and SDAE (AUC: 0.78 to 0.94, AP: 0.66 to 0.86) in distinguishing between coronary cross-sections with mild stenosis (stenosis grade<0.3) and coronary cross-sections with severe stenosis (stenosis grade>0.7) . The improvements were statistically significant (Mann-Whitney U-test, p<0.001).

Similarly, The SDAE-MSR achieved the best aggregated AUC (AP) with an 18% (18%) improvement upon the SAE and SDAE (AUC: 0.71 to 0.84, AP: 0.68 to 0.80). The improvements were statistically significant (Mann-Whitney U-test, p<0.05).

**Conclusion:** Deep sparse autoencoders with mixed structure regularization (MSR) in addition to explicit sparsity regularization term and stochastic corruption of the input data with Gaussian noise have the potential to improve unsupervised abnormality detection using deep-learning compared to common deep autoencoders.

**Introduction:**

Automatic detection of abnormalities in medical images has the potential to provide objective and accurate analysis of medical images, reduce radiologists workload and overall health-related costs.[1,2] Traditionally, discriminative machine-learning approaches were used to design algorithms that are capable to distinguish between normal and abnormal samples.[3–9] In the past few years, discriminative deep-learning approaches demonstrated substantial improvement over traditional machine learning approaches in different tasks including image classification[10] and segmentation.[11] Similarly, such approaches were applied successfully for abnormality detection in medical images.[10,12,13] For a general review of deep-learning approaches for computer aided detection of abnormalities in medical images we refer the reader to the review by Hoo-Chang et al.[14]

However, these discriminative deep-learning approaches rely heavily upon the availability of vast amounts of both normal and abnormal samples with explicit annotation by experts which is very challenging to collect. Moreover, due the large variability of abnormal samples, these models usually require specific training tuned to detect each abnormality. These requirements are identified as significant barriers in developing reliable large-scale models which can be utilized in the clinic. [1,2,14]

Recently, deep autoencoder models were suggested as an alternative to discriminative deep-learning approach for abnormality detection.[15] The autoencoder models learn a representation of normal data by encoding it into a small number of features, and then restore the data using a decoder which takes the encoded features and expand them. The abnormalities are implicitly characterized as outliers from the distribution of normal patients. The main advantage of deep autoencoders is that their training requires normal samples only, without any additional expert annotation of abnormalities. In addition, by modeling abnormalities as outliers the generative models has the potential to cope better with abnormalities heterogeneity compared to the discriminative models.[16]

In their simplest form, autoencoders learn an underlying compact representation by enforcing a bottleneck of a small number of features between the encoder and the decoder components.[17] However, more modern autoencoders employed an overcomplete representation approach in which the autoencoder encode the input data into a rich higher-dimensional representation. In this setup, regularization is required to prevent the autoencoder from converging into representations that are over-fitted to the specific training data.

A common regularization technique is to add a sparsity term to the loss function minimized during the training session (SAE).[17,18] The additional sparsity term explicitly encourage the model to generate representations that are sparse. In contrast, Vincent et al[19] propose the denoising autoencoders (DAE) in which the input data is stochastically corrupted by some type of noise. The DAE implicitly yields a sparse higher level representations through its training criteria: a robust reconstruction of the uncorrupted example. Specifically, they considered three types of noise: 1) Additive isotropic Gaussian noise, 2) Masking noise, and 3) Salt & Pepper noise. However, by corrupting the input data with noise, the DAE may learn how to denoise the input data rather than being more sensitive to the underlying structure of interest.

To overcome this limitation we introduce in this work the "mixed structure regularization" (MSR) approach to regularize deep-learning sparse overcomplete autoencoders. We implicitly encourage the model to learn a sparse representation by stochastically corrupting the input data with randomly added structure sampled form the training data set rather than corrupting with random noise as suggested by Vincent et al.[19]

We evaluate the added value of the proposed MSR approach in unsupervised coronary stenosis detection from Cardiac Computed Tomography Angiography (CCTA) data compared to the sparse DAE (SDAE) and SAE.

## Materials and methods:

### Datasets:

The dataset we used in this study consists of retrospectively collected CCTA data of 90 subjects who underwent a CCTA exam due to suspected CAD. Patients with prior myocardial infarction (MI) or percutaneous coronary intervention (PCI) were excluded from the study. CCTA data was acquired using either Philips Brilliance iCT or Philips Brilliance 64. Acquisition mode was either helical retrospective ECG gating (N=48) or prospectively-gated axial scans ("Step & Shoot Cardiac") (N=42). The kVp range was 80-140 kV and the tube output range was 600-1000 mAs for the helical retrospective scans and 200-300 mAs for the prospectively-gated scans. The voxel size range was: $[028 - 0.48] \times [0.28 - 0.48] \times [0.67 - 0.9] \text{mm}^3$.

### Methods:

**Data preparation**

We computed the coronary artery centerlines and the aorta segmentation automatically with a commercially available software dedicated for cardiac image analysis (Comprehensive Cardiac Analysis, IntelliSpace Portal 6.0, Philips Healthcare). We further manually adjusted the result of the automatic software to improve overall centerlines quality using its graphical user interface.

Next, we segmented the coronary lumen and wall using the automatic coronary lumen segmentation algorithm of Freiman et al[20] with manual adjustments where required.

We then extracted 3D coronary cross-sectional patches of size 80x80x8 mm along the coronary centerline using an in-house software. Inspired by Kitamura et al[21], we calculate the ratio between the wall area ($W_a$) and lumen area ($L_a$) for each coronary cross-section and define a cross-sectional stenosis grade as:

$$[1] \qquad 1 - \frac{L_a}{W_a}.$$

Finally, we select coronary cross-sections with stenosis grade < 0.2 from the training datasets and used them to train the neural networks.

**Deep sparse autoencoder**

The basic autoencoder (AE) is a combination of an encoder and a decoder. A single layer in the encoder is defined as:

$$[2] \qquad h = f(x) = s_f(W_E x + b_E)$$

where $x$ is the input data, $W_E$ represents the convolutional weights, $b_E$ is the bias component, and $s_f$ is a non-linear activation function. A single layer of the decoder is defined as:

$$[3] \qquad y = g(h) = s_g(W_D h + b_D)$$

where $h$ is the output of the encoder, $W_D$ represents the decoder convolutional weights and $b_D$ is the bias component.

A deep autoencoder is built by stacking multiple encoding layers for the encoder and decoding layers for the decoder.

The autoencoder training process consists in finding the set of parameters $\theta = \{W_E, W_D, b_h, b_y\}$ that minimize a reconstruction error on a training set of examples $D_n$. Formally:

[4] $$\hat{\theta} = \arg\min_{\theta} \sum_{x \in D_n} L\left(x, g(f(x))\right)$$

where $L$ is the reconstruction error $\|x - g(f(x))\|$.

Regularized autoencoder[17] avoids overfitting by adding a regularization term that penalizes for large values of $W = \{W_E, W_D\}$:

[5] $$\hat{\theta} = \arg\min_{\theta} \sum_{x \in D_n} L\left(x, g(f(x))\right) + \lambda R(W)$$

where $R$ is a regularization function such as the L1 or the L2 norms and $\lambda$ is a meta-parameter that controls the influence of the regularization.

Ng et al proposed the Sparse autoencoder (SAE)[17,18] which seek to find a sparse representation $h$ in a higher dimensional space, by allowing the number of features in $h$ to be larger than in the input data $x$, but penalizing also for non-sparse representations:

[6] $$\hat{\theta} = \arg\min_{\theta} \sum_{x \in D_n} L\left(x, g(f(x))\right) + \lambda R(W) + \gamma S(f(x))$$

where $S$ is the sparsity penalty and $\gamma$ is a meta-parameter that controls the influence of the sparsity regularization.

Vincent et al[19] proposed the denoising autoencoder (DAE) which implicitly encourage sparse representation by corrupting the input data during the training procedure with some noise. Formally, the training of DAE is defined as:

[7] $$\hat{\theta} = \arg\min_{\theta} \sum_{x \in D_n} L\left(x, g(f(\tilde{x}))\right) + \lambda R(W)$$

where in the case of additive random Gaussian noise:

[8] $$\tilde{x} = x + z_i, \qquad z_i \sim \mathcal{N}(0,1).$$

Similarly to the added noise in DAE we propose the mixed-structure regularization (MSR), in which $\tilde{x}$ is a weighted combination of the original input and other randomly chosen input sample:

[9] $$\tilde{x} = (1-\alpha)x + \alpha x_r$$

where $x_r \in D_n, x_r \neq x$, and $\alpha$ is a meta-parameter controlling the amount of the random structure that is added to the input.

Finally, we can combine the different regularization approaches together in the training process which now will seek to find the network parameters $W$ that minimize the corresponding loss function as follows:

[10] $$\hat{\theta} = \arg\min_{\theta} \sum_{x \in D_n} L\left(x, g\left(f(\tilde{\tilde{x}})\right)\right) + \lambda R(W) + \gamma S(f(x))$$

where:

[11] $$\tilde{\tilde{x}} = (1-\alpha)x + \alpha x_r + z_i, \qquad z_i \sim \mathcal{N}(0, \sigma).$$

The architecture of our deep sparse autoencoder is illustrated in Figure 1. Our deep autoencoder architecture consists of two layers of encoding and two layers of -decoding. Each encoding layer consists of 3x3x3 convolutional layer followed by non-linear activation function of the form of parametric rectified linear unit[22] and max-pooling with sampling factor of 2. Each decoding layer consists of 3x3x3 convolutional layer followed by non-linear activation function of the form of parametric rectified linear unit[22] and upsampling by factor of 2. At the final layer a 1x1x1 convolution is used to map each component feature vector to the output pixel value.

**Abnormality detection**

We define the group of normal samples from the validation set ($S_{normal}$). For each sample in $S_{normal}$, we calculate L1 reconstruction error of the autoencoder:

[12] $$err(x) = |x - g(f(x))|$$

where $x \in S_{normal}$ are the input patches, $g(f(x))$ are are the output patches generated by the auto-encoder.

Finally, we calculate the mean and standard deviation ($\mu_{normal\_err}, \sigma_{normal\_err}$) over all reconstruction errors $err(x), x \in S_{normal}$.

Next, we detect abnormalities by:

1) feeding the network with the patient image patches ($x$),
2) computing the L1 reconstruction error between the input and the output images (Eq. 12)
3) calculating the abnormality grade as the number of pixels with reconstruction error > $\mu_{normal\_err} + 3\,\sigma_{normal\_err}$.

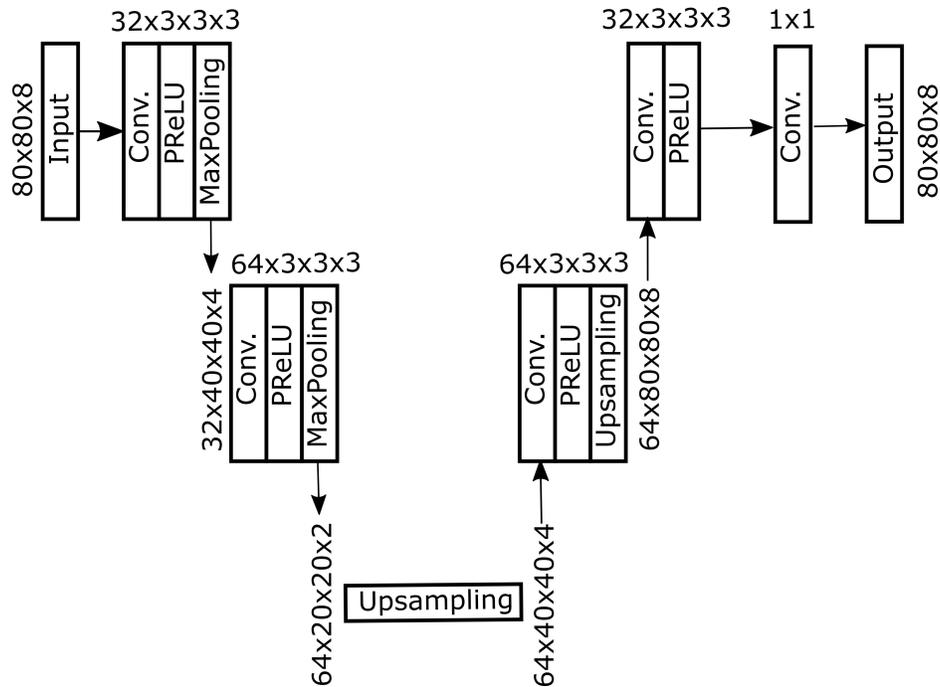

**Fig. 1:** The deep autoencoder architecture. Left side is the encoder and right side is the decoder.

**Evaluation methodology:**

We used 3D image patches of dimension 80x80x8, sampled from the straight multi-planar reformatted CCTA data along the coronary centerline.

We divided the data into ten groups of 9 subjects each. Following a 10-fold cross-validation scheme, we generated 10 datasets, each consists of a training dataset (54 cases), validation dataset (18 cases) and test dataset (18 cases).

We trained the autoencoder using a momentum stochastic gradient decent algorithm with momentum of 0.9.[23] We used a hierarchical training approach, in which the number of minibatches, the number of epochs, and the learning parameters were modified at each stage to accelerate and stabilize the convergence of the training procedure. Minibatch size of 32 samples was used for all stages. Table 1 summarizes the learning parameters.

We optimized the value of the hyper-parameters $\lambda, \gamma, \alpha, \sigma$ using the validation data using a grid-search approach. Table 2 summarizes the values of the hyper-parameters for each autoencoder version that we evaluated.

We first evaluated the added-value of the mixed structure regularization (MSR) and additive Gaussian noise in addition to the explicit sparsity term in the loss function in distinguishing between coronary cross-sections with no or mild stenosis (stenosis grade < 0.3) and coronary cross-sections with severe stenosis (stenosis grade > 0.7) using the test dataset (20 cases) for each fold of the 10-fold cross-validation scheme.

Next, we assessed the performance of the MSR autoencoder in detecting coronary cross-sections with at least intermediate stenosis grade (0.4 and above) using each test dataset for each fold of the 10-fold cross-validation scheme.

We used aggregated Receiver Operation Characteristics (ROC) curves with area under the curve (AUC) and Precision-Recall (PR) curves to analyze the performance of each regularization approach. We also used box-plots to summarize the distribution of the AUC and PR for the different folds along with the Mann-Whitney U-test to assess the statistical significance of the difference between the various regularization approaches.

**Results:**

Table 3 summarizes the distribution of the stenosis grades in our dataset. Fig. 2 presents representative coronary sections from the training dataset (stenosis grade<0.2) with the noise and random noise structure used during the training phase of the different encoders. Fig. 3 depicts representative coronary cross sections with stenosis grades of 0.2, 0.5, and 0.8 along with their reconstruction results using the SDAE-MSR autoencoder.

**Table 1**: Summary of the learning parameters for each stage of the training procedure

| Stage | Number of epochs | Number of minibatches per epoch | Learning rate |
|---|---|---|---|
| **1** | 100 | 100 | 0.001 |
| **2** | 80 | 200 | 0.0005 |
| **3** | 60 | 300 | 0.00025 |
| **4** | 40 | 500 | 0.0001 |

**Table 2**: Hyper-parameter settings for each type of autoencoder

| Autoencoder type | $\lambda$ | $\gamma$ | $\alpha$ | $\sigma$ |
|---|---|---|---|---|
| **SAE** | 0.001 | 0.0005 | 0.0 | 0.0 |
| **SDAE** | 0.001 | 0.0005 | 0.0 | 0.1 |
| **SAE-MSR** | 0.001 | 0.0005 | 0.1 | 0.0 |
| **SDAE-MSR** | 0.001 | 0.0005 | 0.1 | 0.001 |

**Table 3**: Stenosis grades distribution in our database by group (average, [range]).

| | $Total$ | $< 0.2$ | $< 0.3$ | $> 0.7$ | $< 0.4$ | $>= 0.4$ |
|---|---|---|---|---|---|---|
| **Training** | 8892 [8234-9866] | 8892 [8234-9866] | - | - | - | - |
| **Validation** | 7586 [6631-8619] | - | 2817 [2046-3382] | 1132 [858-1405] | 4229 [3268-5211] | 3357 [2881-3936] |
| **Test** | 7586 [6631-8619] | - | 2817 [2046-3382] | 1132 [858-1405] | 4229 [3268-5211] | 3357 [2881-3936] |

Fig. 4 presents straight multi-planar reformatted images of representative coronary arteries at different viewing angles along with the abnormality grade that was given by SDAE-MSR.

Fig. 5 presents the aggregated performance curves for the different autoencoders in distinguishing between coronary cross-sections with mild or below stenosis (stenosis grade<0.3) and coronary cross-sections with severe stenosis (stenosis grade >0.7) along with box-plot representations of the distribution of the AUC and AP for the different folds. The SDAE-MSR achieved the best aggregated AUC (AP) with a 20% (30%) improvement upon the SAE and SDAE (AUC: 0.78 to 0.94, AP: 0.66 to 0.86). The improvements in the AUC (AP) achieved by the SDAE-MSR and SAE-MSR compared to the AUC of the SAE or the SDAE were statistically significant (Mann-Whitney U-test, $p<0.001$). The combination of all sparsity promotion approaches in the SDAE-MSR improved the aggregated AUC from 0.93 to 0.94. However, the difference was not statistically significant. Similarly, the SDAE did not improve AUC and AP upon the SAE.

Fig. 6 presents the aggregated performance curves for the different autoencoders in detecting coronary cross-sections with moderate or above stenosis (stenosis grade>0.4) along with box-plot representations of the distribution of the AUC and AP for the different folds. The SDAE-MSR achieved the best aggregated AUC (AP) with a 18% (18%) improvement upon the SAE and SDAE (AUC: 0.71 to 0.84, AP: 0.68 to 0.80). There was not difference in aggregated AUC and AP between the SDAE-MSR and the SAE-MSR. The improvements in the AUC (AP) achieved by the SDAE-MSR and SAE-MSR compared to the AUC of the SAE or the SDAE were statistically significant (Mann-Whitney U-test, $p<0.05$).

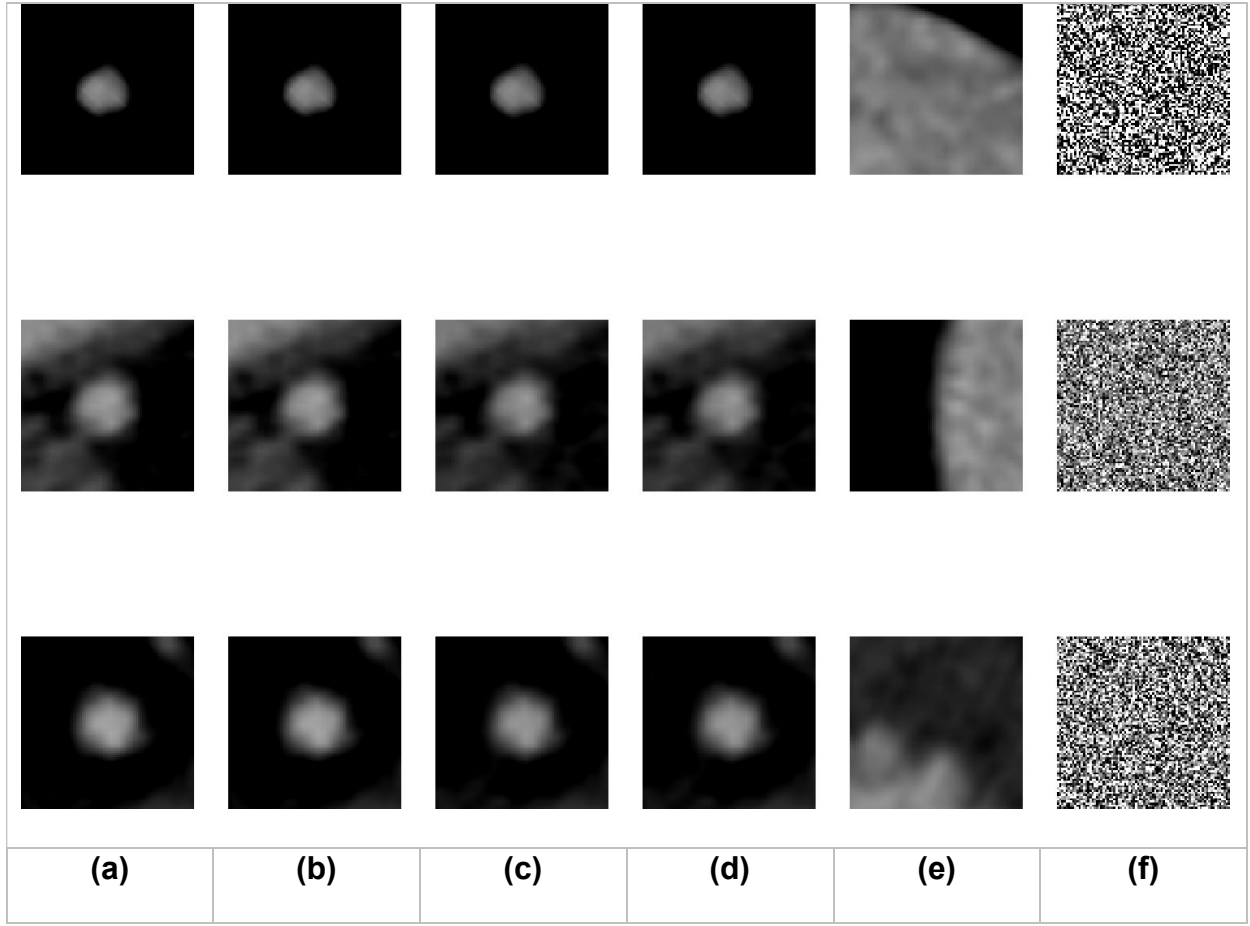

**Fig. 2:** (a) Representative coronary cross-sections from the training set (stenosis grades < 0.2), along with their synthesized versions: (b) added random noise (c) mixed structure, (d) combined mixed structure and random noise, (e) the structure used in the synthesized version, and; (f) the random noise used in the synthesized version. Note that since the weights of the added structure or noise are very small, their effect on the patch is almost non-visible to the human eye.

16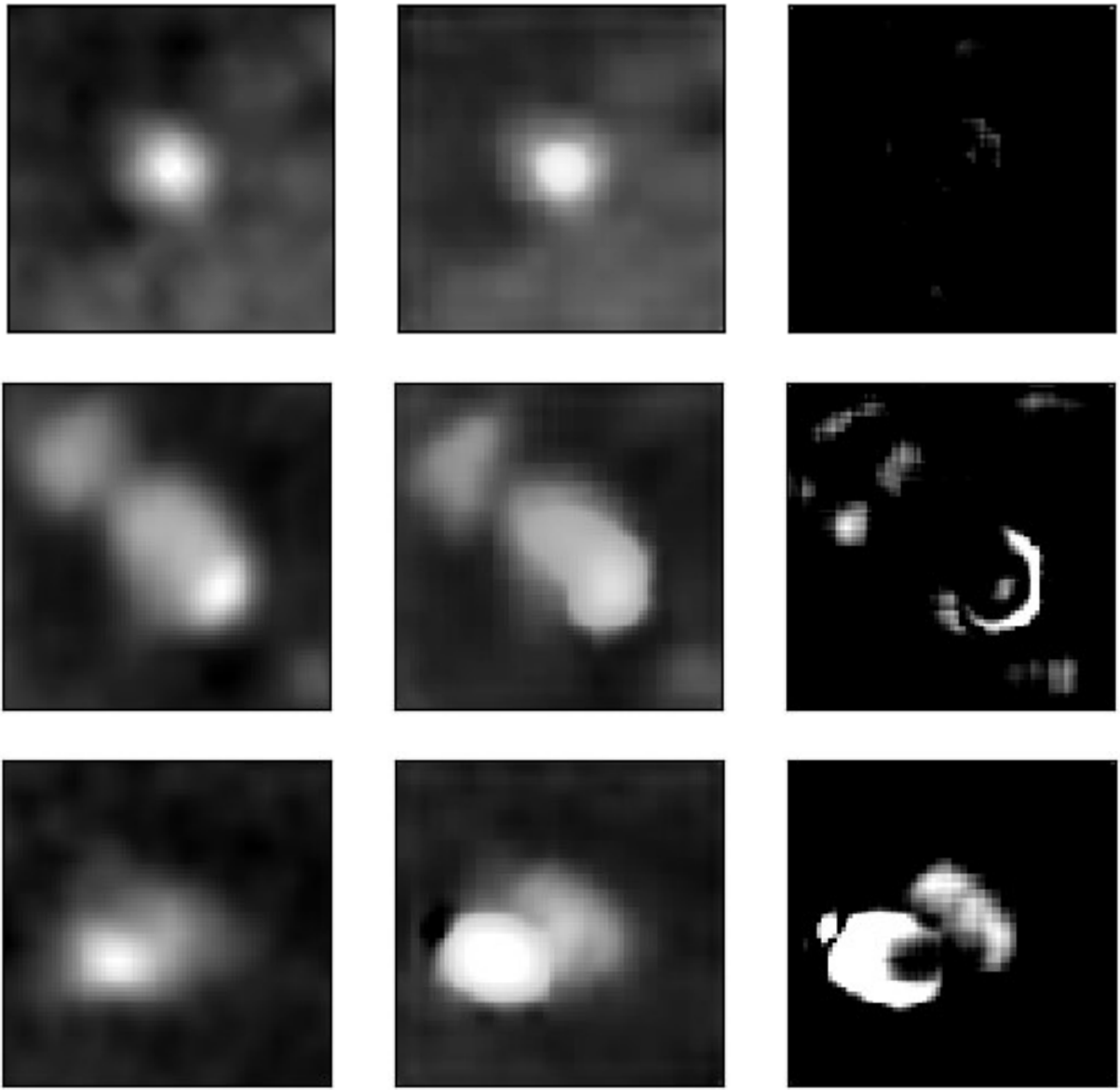

**Fig. 3:** Representative coronary cross-sections (left column) with stenosis grades of 0.2 (top row), 0.5 (middle row) and 0.8 (bottom row) along with their reconstruction results using the SDAE-MSR autoencoder (middle column) and the difference between the input cross-section and the reconstructed cross-section (right column).

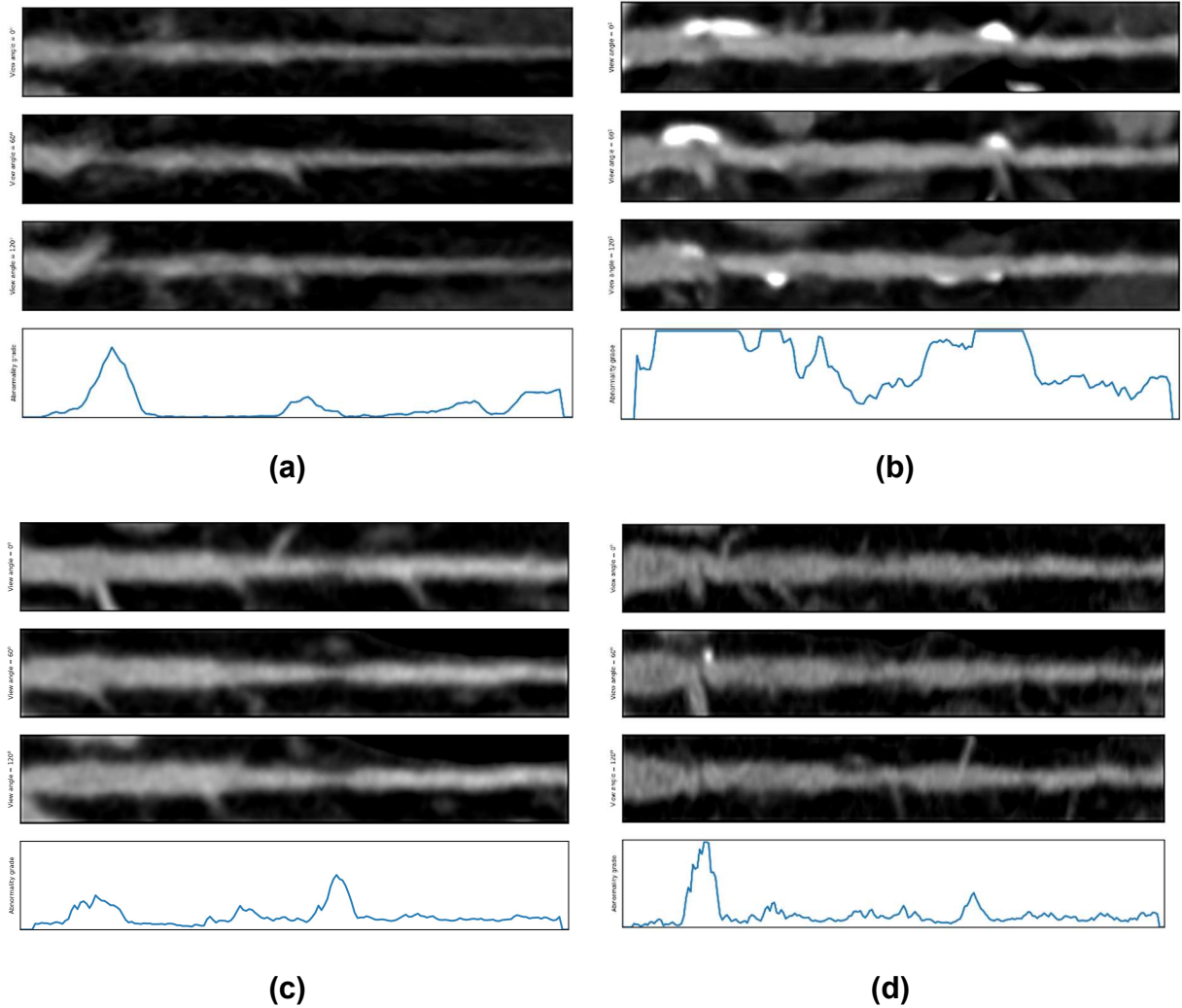

**Fig. 4:** Representative straight multi-planar reformatted coronary arteries at different viewing angles ($0^0$, $60^0$, $120^0$, 1st-3rd rows), along with the abnormality grade given by our SDAE-MSR autoencoder (4th row). High abnormality grade indicate potential stenosis in the coronary.

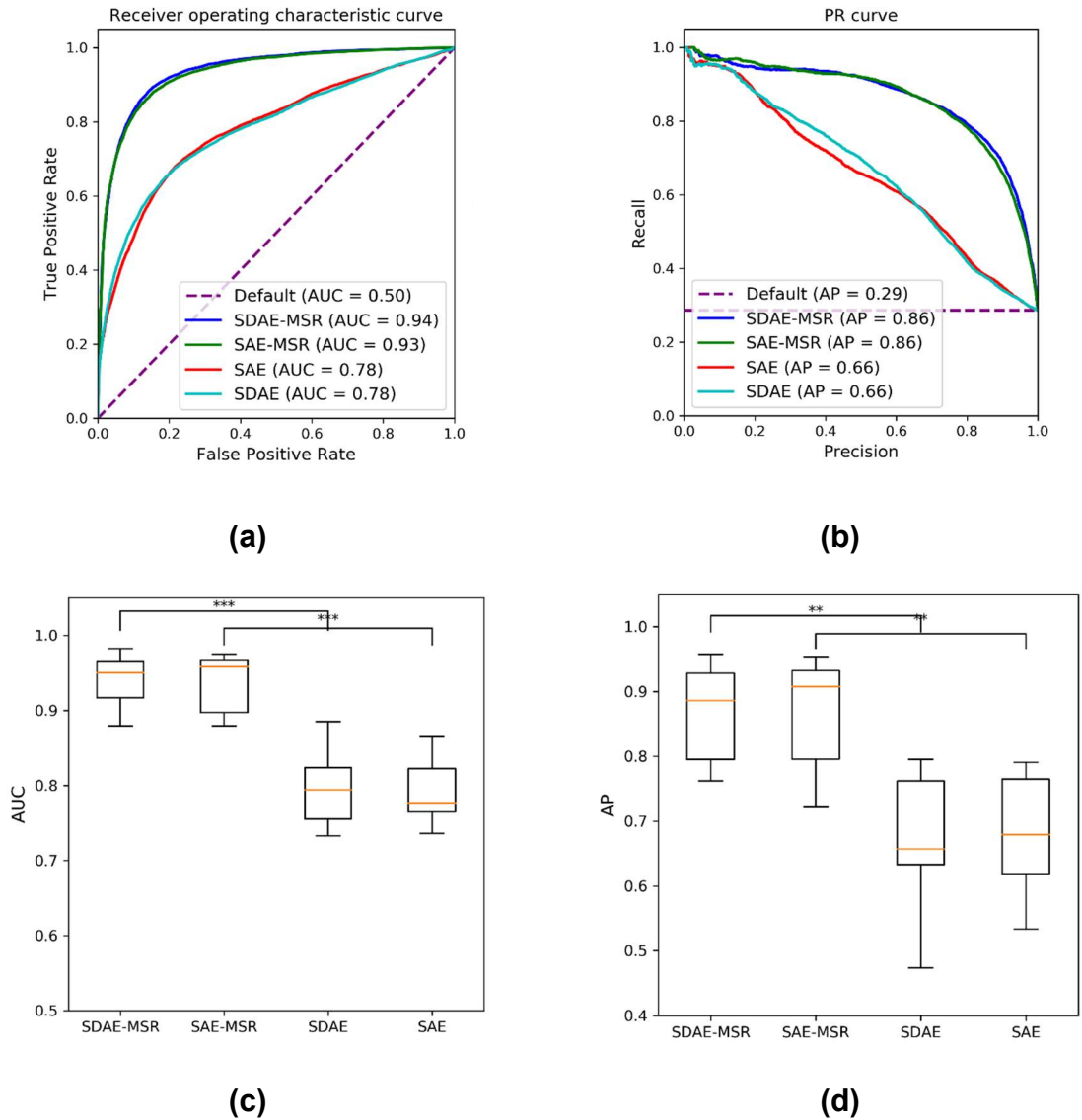

**Fig. 5:** Performance summary for the different autoencoders in distinguishing between coronary cross-section with mild stenosis or below (stenosis grade<0.3), and coronary cross-section with severe stenosis (stenosis grade>0.7). **(a)** aggregated ROC curve, **(b)** aggregated Precision-Recall curve, **(c)** fold-wise AUC, and; **(d)** fold-wise AP.

The MSR improve overall AUC and AP significantly (*** p-value<0.0001, ** p-value<0.001. Added random noise did not show a significant improvement.

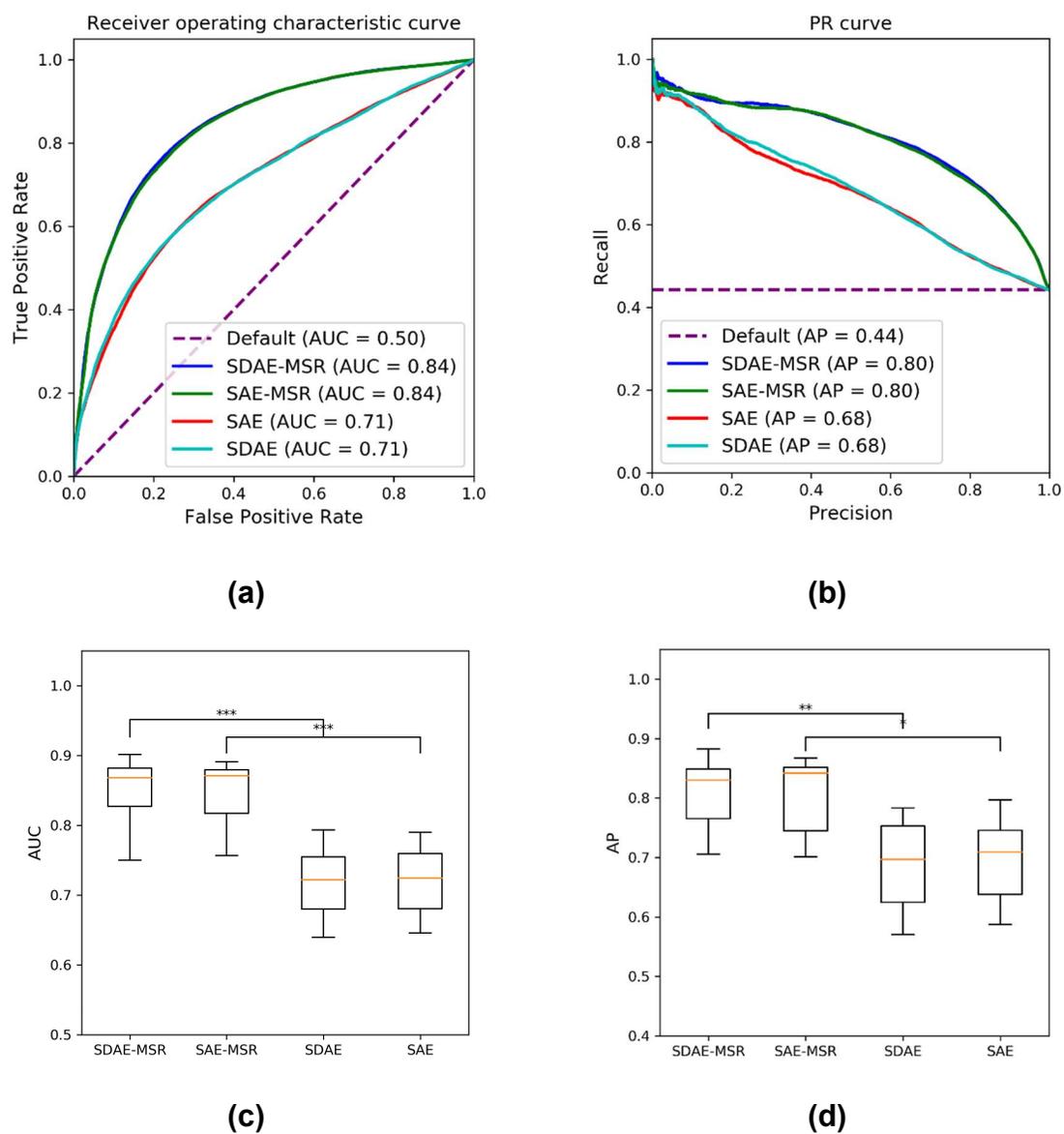

**Fig. 6:** Performance summary for the different autoencoders in detecting coronary cross-section with moderate or above stenosis (stenosis grade>0.4). **(a)** aggregated ROC curve, **(b)** aggregated Precision-Recall curve, **(c)** fold-wise AUC, and; **(d)** fold-wise AP. The MSR improve overall AUC and AP significantly (*** p-value<0.0001, ** p-value<0.001. Added random noise did not show a significant improvement.

**Discussion:**

Our study demonstrates the potential of mixed structure regularization (MSR) in regularizing deep sparse autoencoders used for unsupervised abnormality detection.

Differentiating abnormal appearance of pathologies in medical images from normal appearance by reliably modeling the normal appearance has several advantages over explicit classification of abnormalities in medical images. Examples include: 1) The need for normal data only without any expert annotation on it. 2) Ability to cope with multiple types of abnormalities rather than explicitly model each abnormality. [1,2,14]

Several deep-learning based models have been recently suggested for unsupervised abnormality detection. Schlegl et al.[25] used generative adversarial network (GAN) to detect abnormalities. In the training phase, they trained a GAN to map random samples from a latent space to the normal data. Then, they detect abnormalities by finding the latent representation of the given image. Finally, the distance between the image reconstructed by the GAN and the input image was used to determine the presence of abnormality in the image.

More recently, deep autoencoders were suggested as an efficient tool to model normal appearance in medical images. Sato et al[26] used a spatial autoencoder in which the dimension of the latent space is smaller than the dimension of the input space, thus ensuring compressed representation, to model normal appearance in CT images of the brain.

In contrast, overcomplete autoencoders allow a higher dimensional latent space which has the potential to better represent the input image. Several studies suggested that overcomplete auto-encoders may learn a more generalizable representation of the input and therefore yield improved performance in various tasks. For example, Vincent et al[19] show that overcomplete auto-encoders regularized by noise learn a meaningful hidden representation of images while bottleneck autoencoders were not successful in learning a meaningful underlying representation. More recently, Watkins et al[27] show that overcomplete sparse autoencoders performed better than bottleneck autoencoders for image compression tasks.

However, some form of regularization is required to avoid overfitting of the overcomplete auto-encoder to the training data. Baur et al[15] propose a combination of pixel-wise reconstruction error based on an $L^p$ distance with an explicit sparsity term[17] and an adversarial loss to improve the quality of the reconstructed images during the training phase, therefore obtaining a better representation of the appearance of normal data. Bergmann et al[28] suggest to replace the pixel-wise reconstruction error based on an $L^p$ distance with overall image structural similarity index (SSIM).[29]

Denoising autoencoders (DAE), proposed by Vincent et al[19], stochastically corrupt the input data by some type of noise and aim to minimize a reconstruction error suggest an alternative implicit regularization scheme. However, both explicit sparsity terms and corruption with added random noise may result in overly smoothed reconstructed images which may reduce the capability of the autoencoder to detect abnormalities.

In this work we introduce the 'mixed structure regularization' (MSR), in which the original input image is corrupted by additional random structure, sampled randomly from the training data. The image corruption by additional structure encourages the autoencoder to be more robust to image variability while avoiding over-smoothing of unrelated noise.

We demonstrated the added value of the MSR in the specific task of unsupervised detection of coronary stenosis from CCTA data. Our experiments show that while SDAE and SAE performed about the same in both distinguishing between coronary cross-sections with mild stenosis (stenosis grade<0.3) and coronary cross-sections with severe stenosis (stenosis grade>0.7) and in detecting coronary cross-sections with moderate or above stenosis (stenosis grade>0.4), adding the MSR improved the performance of the SAE (SAE-MSR). Moreover, by corrupting the input with both MSR and random Gaussian noise (SDAE-MSR), we further improved the performance of the AE in both tasks.

These results suggest that MSR may have the potential to further improve the ability of deep sparse autoencoder to reliably learn sparse representations and to improve AE performance in various applications.

Several studies suggested to use the features generated by auto-encoders as input in the setting of supervised classification. For example, Chen et al[12,13] suggested to use features generated by stacked denoising auto-encoder as input to a multi-task regression system in order to classify lung pulmonary nodules from CT data. Future extensions of the MSR can include utilized MSR based autoencoders as features generators for subsequent supervised classification systems.

Our study has several limitations: First, we evaluated the added value of the mixed structure regularization on the context of unsupervised abnormality detection, and more specifically on the task of unsupervised coronary stenosis detection from CCTA. While our result show the added value of the MSR, there is a room to explore the added value of the MSR in other tasks in which deep autoencoders are applied to.

Second, we focused our experiments on demonstrating the added value of the MSR in regularizing deep sparse autoencoder for unsupervised abnormality detection. A more comprehensive system with additional components such as data augmentation, adversarial training[25], and supervised training[12,13] may be used in practice to achieve higher accuracy in the abnormality detection task.

Finally, we defined the reference stenosis grading for the experiments based on the results of an automatic segmentation of the coronary lumen and wall in order to facilitate the generation of large amounts of data as required for deep neural networks training and reliable evaluation. Although the segmentation results reviewed by expert to ensure correctness, there is a room for more accurate data with a detailed expert stenosis grading in a clinical setting to improve the overall performance of the abnormality detection algorithm.

In conclusion, we have presented the mixed structure regularization approach for regularizing deep sparse autoencoders. Deep sparse autoencoders have the potential to enable automatic medical abnormality detection by training solely on normal data. We demonstrated the added value of mixed structure regularization in addition to other techniques including explicit sparsity term, and denoising autoencoder in detecting coronary stenosis from CCTA data. The proposed regularization approach has the potential to improve the performance of deep sparse autoencoders.

**Disclosure of Conflicts of Interest:**

M.F. and L.G. are employees of Philips Healthcare.

**References:**

1. van Ginneken B, Schaefer-Prokop CM, Prokop M. Computer-aided Diagnosis: How to Move from the Laboratory to the Clinic. *Radiology*. 2011. doi:10.1148/radiol.11091710

2. Wang S, Summers RM. Machine learning and radiology. *Med Image Anal*. 2012. doi:10.1016/j.media.2012.02.005

3. Cortes C, Vapnik V. Support-Vector Networks. *Mach Learn*. 1995. doi:10.1023/A:1022627411411

4. Larroza A, Moratal D, Paredes-Sánchez A, et al. Support vector machine classification of brain metastasis and radiation necrosis based on texture analysis in MRI. *J Magn Reson Imaging*. 2015. doi:10.1002/jmri.24913

5. Sela Y, Freiman M, Dery E, et al. FMRI-based hierarchical SVM model for the classification and grading of liver fibrosis. *IEEE Trans Biomed Eng*. 2011;58(9):2574-2581.

6. Setio AAA, Jacobs C, Gelderblom J, Van Ginneken B. Automatic detection of large pulmonary solid nodules in thoracic CT images. *Med Phys*. 2015. doi:10.1118/1.4929562

7. Breiman L. Random forests. *Mach Learn*. 2001. doi:10.1023/A:1010933404324

8. Huang P, Park S, Yan R, et al. Added Value of Computer-aided CT Image Features for Early Lung Cancer Diagnosis with Small Pulmonary Nodules: A Matched Case-Control Study. *Radiology*. 2018. doi:10.1148/radiol.2017162725

9. Lebedev A V., Westman E, Van Westen GJP, et al. Random Forest ensembles for detection and prediction of Alzheimer's disease with a good between-cohort

robustness. *NeuroImage Clin*. 2014. doi:10.1016/j.nicl.2014.08.023

10. Krizhevsky A, Sutskever I, Hinton GE. ImageNet Classification with Deep Convolutional Neural Networks. *Adv Neural Inf Process Syst*. 2012. doi:http://dx.doi.org/10.1016/j.protcy.2014.09.007

11. Chen LC, Papandreou G, Kokkinos I, Murphy K, Yuille AL. DeepLab: Semantic Image Segmentation with Deep Convolutional Nets, Atrous Convolution, and Fully Connected CRFs. *IEEE Trans Pattern Anal Mach Intell*. 2018. doi:10.1109/TPAMI.2017.2699184

12. Chen S, Qin J, Ji X, et al. Automatic Scoring of Multiple Semantic Attributes with Multi-Task Feature Leverage: A Study on Pulmonary Nodules in CT Images. *IEEE Trans Med Imaging*. 2017. doi:10.1109/TMI.2016.2629462

13. Chen S, Ni D, Qin J, Lei B, Wang T, Cheng JZ. Bridging computational features toward multiple semantic features with multi-task regression: A study of CT pulmonary nodules. In: *Lecture Notes in Computer Science (Including Subseries Lecture Notes in Artificial Intelligence and Lecture Notes in Bioinformatics)*. ; 2016. doi:10.1007/978-3-319-46723-8_7

14. Shin HC, Roth HR, Gao M, et al. Deep Convolutional Neural Networks for Computer-Aided Detection: CNN Architectures, Dataset Characteristics and Transfer Learning. *IEEE Trans Med Imaging*. 2016. doi:10.1109/TMI.2016.2528162

15. Baur C, Wiestler B, Albarqouni S, Navab N. Deep Autoencoding Models for Unsupervised Anomaly Segmentation in Brain {MR} Images. *CoRR*. 2018;abs/1804.0. http://arxiv.org/abs/1804.04488.

16. Chandola V, Banerjee A, Kumar V. Anomaly detection: A survey. *ACM Comput Surv*. 2009. doi:10.1145/1541880.1541882

17. Ng A. *Sparse Autoencoder*.

18. Lee H, Ng AY. Sparse deep belief net model for visual area V2. In: *Advances in Neural Information Processing Systems 20*. ; 2008. doi:10.1.1.120.9887

19. Vincent PASCALVINCENT P, Larochelle LAROCHEH H. Stacked Denoising Autoencoders: Learning Useful Representations in a Deep Network with a Local Denoising Criterion Pierre-Antoine Manzagol. *J Mach Learn Res*. 2010. doi:10.1111/1467-8535.00290

20. Freiman M, Nickisch H, Prevrhal S, et al. Improving CCTA-based lesions' hemodynamic significance assessment by accounting for partial volume modeling in automatic coronary lumen segmentation. *Med Phys*. 2017;44(3):1040-1049. doi:10.1002/mp.12121

21. Kitamura Y, Li Y, Ito W, Ishikawa H. Coronary lumen and plaque segmentation from CTA using higher-order shape prior. In: *Lecture Notes in Computer Science (Including Subseries Lecture Notes in Artificial Intelligence and Lecture Notes in Bioinformatics)*. ; 2014. doi:10.1007/978-3-319-10404-1_43

22. He K, Zhang X, Ren S, Sun J. Delving deep into rectifiers: Surpassing human-level performance on imagenet classification. In: *Proceedings of the IEEE International Conference on Computer Vision*. ; 2015. doi:10.1109/ICCV.2015.123

23. Qian N. On the momentum term in gradient descent learning algorithms. *Neural Networks*. 1999. doi:10.1016/S0893-6080(98)00116-6

24. Delong ER, DeLong DM, Clarke-Pearson DL. Comparing the areas under two or

more correlated receiver operating characteristic curves: a nonparametric approach. *Biometrics*. 1988;44(3):837-845. doi:10.2307/2531595

25. Schlegl T, Seeböck P, Waldstein SM, Schmidt-Erfurth U, Langs G. Unsupervised anomaly detection with generative adversarial networks to guide marker discovery. In: *Lecture Notes in Computer Science (Including Subseries Lecture Notes in Artificial Intelligence and Lecture Notes in Bioinformatics)*. ; 2017. doi:10.1007/978-3-319-59050-9_12

26. Sato D, Hanaoka S, Nomura Y, et al. A primitive study on unsupervised anomaly detection with an autoencoder in emergency head CT volumes. In: Vol 10575. ; 2018:105751P-10575-6. https://doi.org/10.1117/12.2292276.

27. Watkins Y, Sayeh M, Iaroshenko O, Kenyon GT. Image Compression: Sparse Coding vs. Bottleneck Autoencoders. *CoRR*. 2017;abs/1710.0. http://arxiv.org/abs/1710.09926.

28. Bergmann P, Löwe S, Fauser M, Sattlegger D, Steger C. Improving Unsupervised Defect Segmentation by Applying Structural Similarity to Autoencoders. *arXiv Prepr arXiv180702011*. 2018.

29. Wang Z, Bovik AC, Sheikh HR, Simoncelli EP. Image quality assessment: From error visibility to structural similarity. *IEEE Trans Image Process*. 2004. doi:10.1109/TIP.2003.819861